%% file: main.tex
\documentclass[11pt]{article}

\usepackage[final]{acl}
\usepackage{wrapfig}
\usepackage[absolute,overlay]{textpos}

\usepackage{times}
\usepackage{subcaption}
\usepackage{latexsym}
\usepackage{multicol}
\usepackage{multirow}
\usepackage[T1]{fontenc}
\usepackage{xspace}

\usepackage[utf8]{inputenc}

\usepackage{microtype}

\usepackage{inconsolata}

\usepackage{graphicx}
\usepackage[absolute,overlay]{textpos}

\newcommand{\github}{\raisebox{-1.5pt}{\includegraphics[height=1.05em]{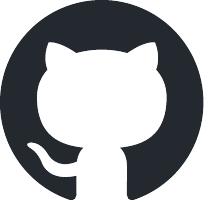}}\xspace}
\newcommand{\huggingface}{\raisebox{-1.5pt}{\includegraphics[height=1.05em]{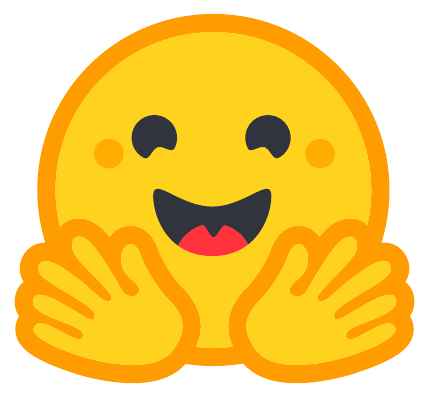}}\xspace}

\usepackage{amsmath}
\usepackage{listings}
\usepackage{float}  

\usepackage[breakable]{tcolorbox} 
\usepackage{xcolor}
\tcbuselibrary{skins}
\tcbuselibrary{raster}
\usepackage{graphicx}
\usepackage{booktabs}
\usepackage{colortbl}
\usepackage{dblfloatfix}

\newcommand{\hide}[1]{} 
\definecolor{prompt-color}{HTML}{B0B0B0}
\definecolor{chosen-color}{HTML}{808080}
\definecolor{rejected-color}{HTML}{808080}
\definecolor{lightblue}{RGB}{80,120,130}
\definecolor{mintgreen}{RGB}{60,180,60}
\definecolor{lavender}{RGB}{120,100,180}

\title{SWE-Dev: Building Software Engineering Agents with Training and Inference Scaling}

\author{Haoran Wang$^{1*}$, Zhenyu Hou$^{1*}$, Yao Wei$^{2}$, Jie Tang$^{1}$, Yuxiao Dong$^1$\\ \\
\textsuperscript{1}Tsinghua University \quad
\textsuperscript{2}Zhipu AI\\[2ex]
\begin{minipage}{\textwidth}
\centering
{\github\ \href{https://github.com/THUDM/SWE-Dev}{\text{Code}} \quad
\huggingface\ \href{https://huggingface.co/THUDM/SWE-Dev-32B}{\text{Models \& Data}}}
\end{minipage}
}

\begin{document}
\maketitle

\renewcommand{\thefootnote}{\fnsymbol{footnote}}
    \footnotetext[1]{Work partially done when interned at Zhipu AI.}
\renewcommand{\thefootnote}{\arabic{footnote}}

\input{paras/abstract}
\input{paras/intro}
\input{paras/related_works}
\input{paras/method}
\input{paras/experiments}

\input{paras/scaling}
\input{paras/conclusion}
\input{paras/limitation}

\paragraph{Acknowledgment.}
{
This work has been supported by the Natural Science Foundation of China (NSFC) 62276148, Tsinghua University (Department of Computer Science and Technology) -Siemens Ltd., China Joint Research Center for Industrial Intelligence and Internet of Things (JCIIOT), and the New Cornerstone Science Foundation through the XPLORER PRIZE.
The GPU compute used in this work is sponsored by Zhipu AI. 
The corresponding author: Yuxiao Dong (yuxiaod@tsinghua.edu.cn). 
}

\bibliography{ref}

\clearpage

\input{paras/appendix}

\end{document}

%% file: paras/abstract.tex
\begin{abstract}

Large language models (LLMs) have advanced rapidly from conversational problem solving to addressing real-world tasks involving tool use, such as software engineering (SWE). 
Recent LLM-powered SWE systems, such as OpenAI Codex and Cursor, have offered end-to-end automation of the software development process. 
However, building effective SWE agents remains challenging due to the lack of high-quality training data and reliable test-time evaluation. 
To address this issue, we present \textsc{SWE-Dev}, an SWE agent built upon open-source LLMs, with a focus on training and inference scaling.
For training scaling, we develop a robust pipeline to synthesize test cases and scale up agent trajectories to construct the training data.
For inference scaling, we increase the interaction budget within a single run to enable further thinking within one independent attempt.
Experiments on the SWE-bench-Verified benchmark show that the \textsc{SWE-Dev} models can achieve top performance among all open SWE agents. 
Specifically, the resolve rate of our 7B and 32B models reach 23.4\% and 36.6\%, respectively, outperforming state-of-the-art open-source models.
All code, models, and datasets are publicly available at \url{https://github.com/THUDM/SWE-Dev}.

\end{abstract}


\hide{Large language models (LLMs) have advanced rapidly from conversational problem solving to addressing real-world tasks with tool use, such as software engineering (SWE). 
Recent LLM-powered SWE toolkits, such as OpenAI Codex and Cursor, have offered end-to-end automation of the software development process. 
However, building effective SWE agents remains challenging due to the lack of high-quality training data and effective test cases. 
To address this issue, we present \textsc{SWE-Dev}, an SWE agent built upon open-source LLMs. 
First, we develop a robust pipeline to synthesize test cases for patch evaluation. 
Second, we scale up agent trajectories to construct the training data for building SWE-Dev. 
Experiments on the SWE-bench-Verified benchmark show that the SWE-Dev models can achieve top performance among all open SWE agents. 
Specifically, the resolve rate of the SWE-Dev 7B and 32B parameter models reach 23.4\% and 36.6\%, respectively, outperforming state-of-the-art open-source models. 
All code, models, and datasets are publicly available at \url{https://github.com/THUDM/SWE-Dev}.}

\hide{
Large language models (LLMs) have advanced rapidly from conversation-based problem solving like competitive math to real-world tasks involving tool calling, such as software engineering. AI-powered toolkits, such as Cursor and OpenAI Codex, offer end-to-end automation of the software development process. However, the development of effective SWE agents is hindered by the scarcity of high-quality training data, primarily due to the scarcity of effective test cases. To address this issue, we present \textsc{SWE-Dev}, an open-source SWE agent with a robust test case generation pipeline. We evaluated our models using the SWE-bench-Verified benchmark and achieved state-of-the-art performance among other open-source SWE agents. Our resolve rate are 23.4\% and 36.6\% for 7B and 32B parameter models, respectively, representing over a 10\% improvement compared to prior work using the same framework. All code, models, and datasets will be released to facilitate future research in the field.

}


%% file: paras/intro.tex
\section{Introduction}

Large language models (LLMs) have rapidly evolved from generating simple code snippets to tackling more complex tasks, such as competitive programming~\citep{Li_2022,openai2025competitiveprogramminglargereasoning}, machine learning problems~\citep{chan2024mlebenchevaluatingmachinelearning}, and real-world software engineering (SWE) tasks~\citep{xi2024agentgymevolvinglargelanguage,jimenez2024swebenchlanguagemodelsresolve,zan2025multiswebenchmultilingualbenchmarkissue,lyu2023enhancing}. 
Among these tasks, SWE is particularly hard and challenging~\citep{openai2025codex, cursor2024}, but highly useful for improving real-world productivity. 
Unlike simple code generation, SWE tasks usually require LLMs to interact with complex and fragile runtime environments, solve toolchain issues, execute scripts, and reason over large, interdependent codebases~\citep{ma2024understandsoftwarerepository}.

\begin{figure}[t]
  \centering
  \vspace{0.5em}
  \includegraphics[width=0.95\columnwidth]{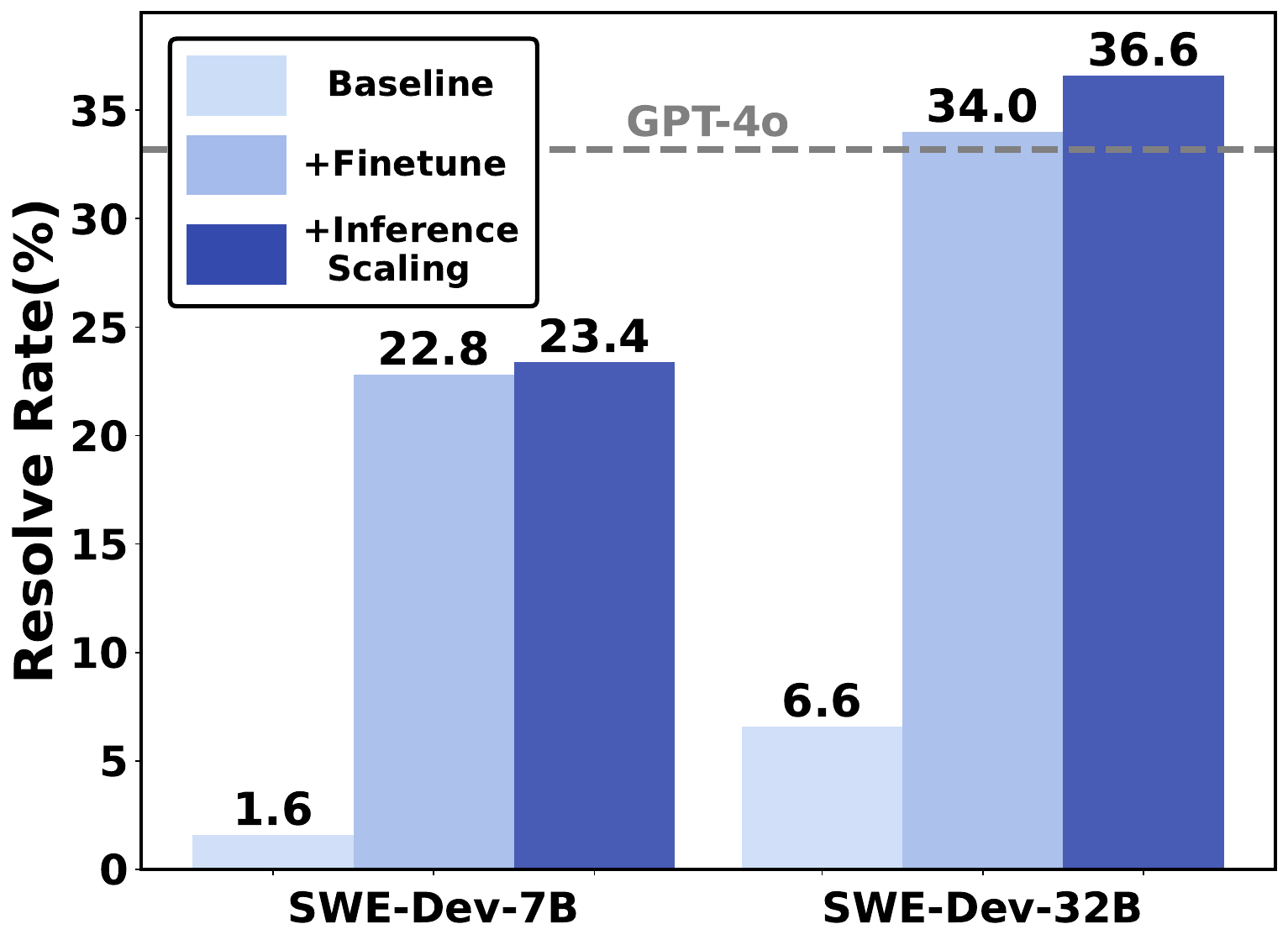}
  \vspace{-0.5em}
  \caption{The \textsc{SWE-Dev} performance with training and inference scaling. Notably, SWE-Dev-32B achieves a resolve rate of 34.0\%, matching the performance of GPT-4o even without  inference scaling.}
  \vspace{-1.5em}
  \label{fig:header}
\end{figure}

The SWE tasks are usually evaluated on benchmarks such as  
SWE-bench~\citep{jimenez2024swebenchlanguagemodelsresolve} and its recent multimodal extensions~\citep{yang2024swebenchmultimodalaisystems, zan2025multiswebenchmultilingualbenchmarkissue}. 
These benchmarks require models to generate verifiable, test-passing solutions on real-world codebases. 
To achieve this, the models must be capable of step-by-step reasoning, tool using, and long-horizon planning.

To date, training models to handle these tasks requires reliable reward signals, typically derived from test cases that validate the correctness of the solutions. 
However, most existing datasets lack such test cases or executable environments~\citep{chen2021evaluatinglargelanguagemodels,lyu2024adapting}, making it difficult to evaluate solutions or provide useful feedback during training. 
This deficiency limits models’ abilities to iteratively refine their outputs through trial-and-error, thus constraining their potential to solve real-world SWE tasks that need grounded, verifiable solutions.

To address this issue, we introduce \textsc{SWE-Dev}, an open-source SWE agent framework coupled with a scalable test case generation pipeline. 
This pipeline works in two-stages: 
First, the LLMs are used to generate Gherkin-style descriptions---a structured natural language format used to specify test scenarios. 
Second, a code generator outputs code patches for further validation. 
Empirical analysis shows that these synthesized test cases align closely with the original problem semantics.

Through large-scale experiments, we identify a clear training scaling trend: increasing the number of sampled agent trajectories leads to improved downstream performance. To enhance efficiency, we propose an LLM-based filtering method that selects high-quality trajectories. This allows us to maintain the benefits of a full dataset while retaining only the most valuable data.

We also propose iteration scaling---a simple yet effective strategy that aims to scale up inference budget by increasing the number of interaction rounds during a single evaluation episode.
This reduces the need for repeated re-evaluation, making it particularly advantageous in scenarios where access to test oracles is expensive or delayed.

In addition, we explore advanced post-training strategies, including 
Rejection Sampling Fine-Tuning (RFT)~\citep{yuan2023scalingrelationshiplearningmathematical}, 
KTO~\citep{ethayarajh2024ktomodelalignmentprospect}, 
and OREO~\citep{wang2024offlinereinforcementlearningllm}. 
We observe that RFT brings the most significant performance improvement, while offline reinforcement learning (RL) methods---KTO and OREO---deliver marginal or task-specific gains.

We build \textsc{SWE-Dev} based on the open Qwen2.5-Coder~\citep{hui2024qwen25codertechnicalreport},
Llama-3.1~\citep{grattafiori2024Llama3herdmodels},  
and GLM4-9B ~\citep{glm2024chatglmfamilylargelanguage} models.  
Its performance is evaluated on the SWE-bench-Verified benchmark, with results shown in Figure~\ref{fig:header}. 
With the Qwen2.5-Coder-32B model, SWE-Dev achieves a resolve rate of 36.6\% on SWE-Bench-Verified, representing state-of-the-art performance among open-source SWE agents. 


\begin{enumerate}
    \item \textbf{Test Case Generation Pipeline.} 
    We build a scalable LLM-based pipeline to generate real-world SWE instances and their executable test cases. 
    Through this pipeline, we successfully construct 2,000 test cases by filtering 38,000 high-quality issues across 4,000 repositories.
    
    \item \textbf{Scaling Trends in Data and Inference.} 
    We empirically identify scaling trends between training data volume, number of interaction steps, and model performance. 
    We find that the SWE-Dev-32B model improves from 34.0\% to 36.6\% resolve rate by adding 45 more interaction turns, highlighting the benefit of multi-step execution for agents.
    
    \item \textbf{Post-Training Recipe for SWE Agents.} 
    We examine several post-training methods, including RFT, KTO, and OREO, as well as hybrid combinations. 
    RFT consistently outperforms others by effectively leveraging high-quality samples, demonstrating its robustness and scalability for training SWE agents.

\end{enumerate}

\hide{

\section{Introduction}

Large language models (LLMs) have rapidly evolved from generating code snippets—as evaluated by benchmarks such as HumanEval~\citep{chen2021evaluatinglargelanguagemodels} and MBPP~\citep{austin2021programsynthesislargelanguage}—to tackling more complex domains, including competitive programming~\citep{Li_2022,openai2025competitiveprogramminglargereasoning}, machine learning tasks~\citep{chan2024mlebenchevaluatingmachinelearning}, and real-world software engineering (SWE) problems such as resolving GitHub issues. This progression is exemplified by benchmarks like AgentGym~\citep{xi2024agentgymevolvinglargelanguage}, SWE-bench~\citep{jimenez2024swebenchlanguagemodelsresolve}, and its multimodal extension~\citep{yang2024swebenchmultimodalaisystems, zan2025multiswebenchmultilingualbenchmarkissue}, which require models to generate verifiable, test-passing solutions within complex, real-world codebases.

Among these tasks, SWE poses particularly formidable challenges~\citep{openai2025codex, cursor2024}. Unlike static code generation, SWE tasks require models to interact with complex and fragile runtime environments, resolve toolchain issues, execute scripts and reason over large, interdependent codebases~\citep{ma2024understandsoftwarerepository}. Successfully solving such problems demands iterative reasoning, tool use, and long-horizon planning.

However, training models to acquire these capabilities hinges on the availability of reliable reward signals—typically derived from test cases that validate the correctness of proposed solutions. Yet existing datasets often lack such test cases or executable environments~\citep{chen2021evaluatinglargelanguagemodels}, making it difficult to evaluate outputs or provide valuable feedback during training. This deficiency limits models’ ability to iteratively refine their outputs or learn from trial-and-error, ultimately constraining performance on real-world SWE tasks that demand grounded, verifiable solutions.

\begin{figure}[t]
  \centering
  \vspace{0.5em}
  \includegraphics[width=0.95\columnwidth]{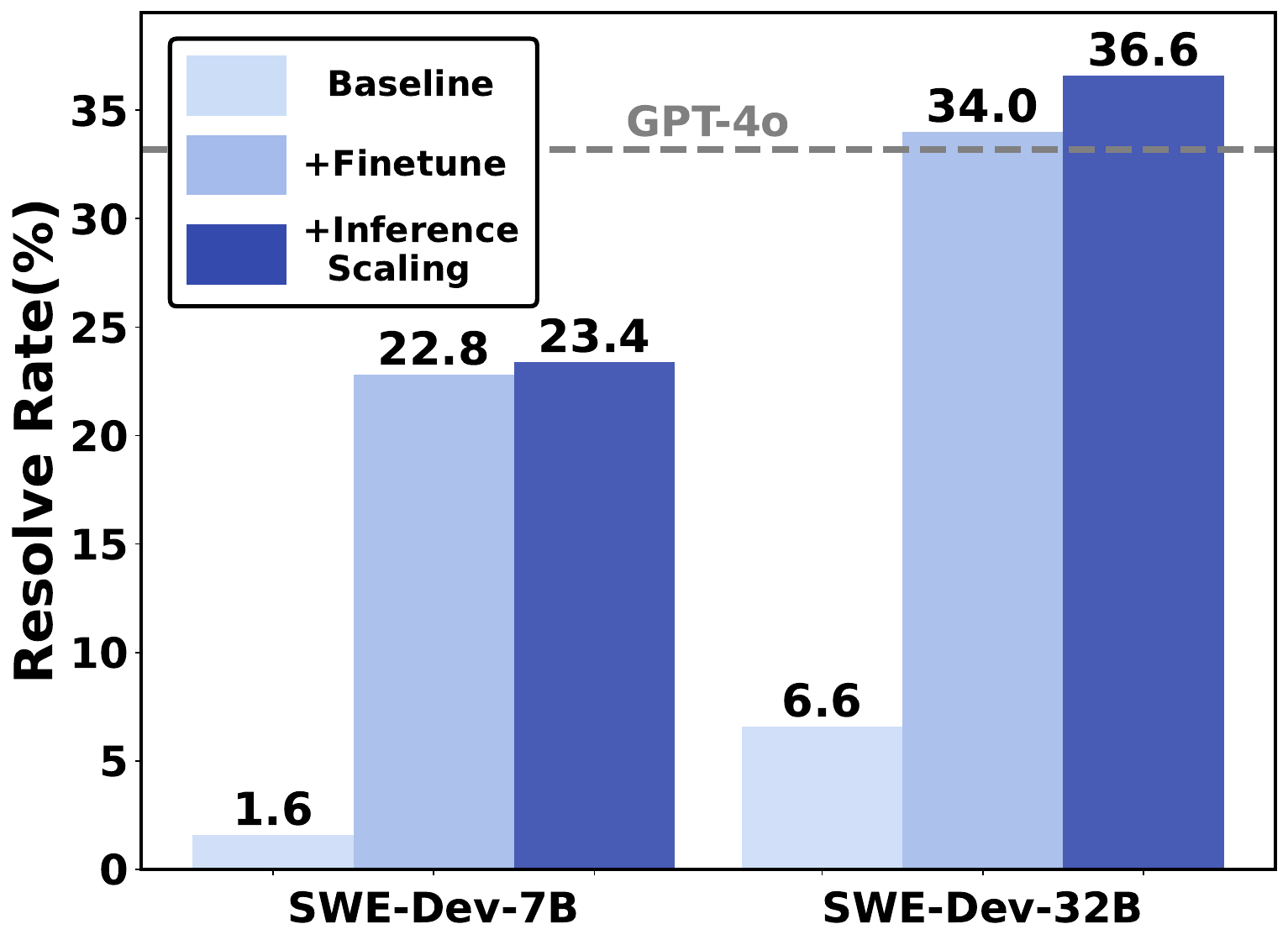}
  \vspace{-0.5em}
  \caption{Model performance with training and inference scaling. Notably, SWE-Dev-32B achieved a performance of 34.0\%, comparable to GPT-4o, even without the benefits of inference scaling.}
  \vspace{-1.5em}
  \label{fig:header}
\end{figure}

To address these limitations, we introduce \textsc{SWE-Dev}, an open-source SWE agent framework paired with a scalable test case synthesis pipeline. This pipeline generates test cases via a two-stage process: first, LLMs are used to produce Gherkin-style descriptions—a structured natural language format for specifying test scenarios, and the code generator will generate corresponding code patches for further validation. Empirical analysis shows that the synthesized test cases align closely with the original problem semantics.

Through large-scale experimentation, we identify a scaling trends between the number of sampled trajectories and downstream performance. To reduce annotation costs, we further propose an LLM-based filtering mechanism that effectively selects high-quality trajectories, substantially compressing dataset size without degrading model performance, and thus preserving learning efficacy.

We also explore advanced post-training methods, including Rejection Sampling Fine-Tuning (RFT)~\citep{yuan2023scalingrelationshiplearningmathematical}, as well as two offline reinforcement learning techniques: KTO~\citep{ethayarajh2024ktomodelalignmentprospect} and OREO~\citep{wang2024offlinereinforcementlearningllm}. Among these, RFT yields the most significant performance improvements, while offline RL methods offer marginal or task-specific gains. Additionally, within the agent-based inference framework, we propose iteration scaling—a simple yet effective strategy that increases the number of interaction rounds during inference. This method reduces the need for repeated re-evaluation, making it particularly advantageous in scenarios where access to test oracles is expensive or delayed.

We evaluate \textsc{SWE-Dev} on the SWE-bench-Verified benchmark, achieving 23.4\% and 36.6\% resolve rate with 7B and 32B models respectively—representing state-of-the-art performance among open-source SWE agents. All code, models, and data will be released for future research.

Our contributions are summarized as follows:

\begin{enumerate}
    \item \textbf{Test Case Generation Pipeline.} We propose a scalable LLM-based pipeline for generating real-world SWE instances and their executable test cases.Through this pipeline, we successfully construct 2,097 test cases by filtering 26k high-quality issues across 4,000 repositories.
    \item \textbf{Scaling Trends in Data and Inference.} We identify empirical scaling trends relating training data volume, number of interaction rounds, and model performance. Notably, the SWE-Dev-32B model achieves 36.6\% instead of 34.0\% with 45 more turns, highlighting the benefit of multi-step execution for agents.
    \item \textbf{Post-Training Recipe for SWE Agents.} We compare several post-training strategies, including RFT, KTO, and OREO, as well as hybrid combinations. RFT consistently outperforms others by effectively leveraging high-quality samples, demonstrating its robustness and scalability for SWE agent training.

\end{enumerate}

}

%% file: paras/related_works.tex
\begin{figure*}[htbp]
  \centering
  \includegraphics[width=0.96\linewidth, trim=0 60 0 50, clip]{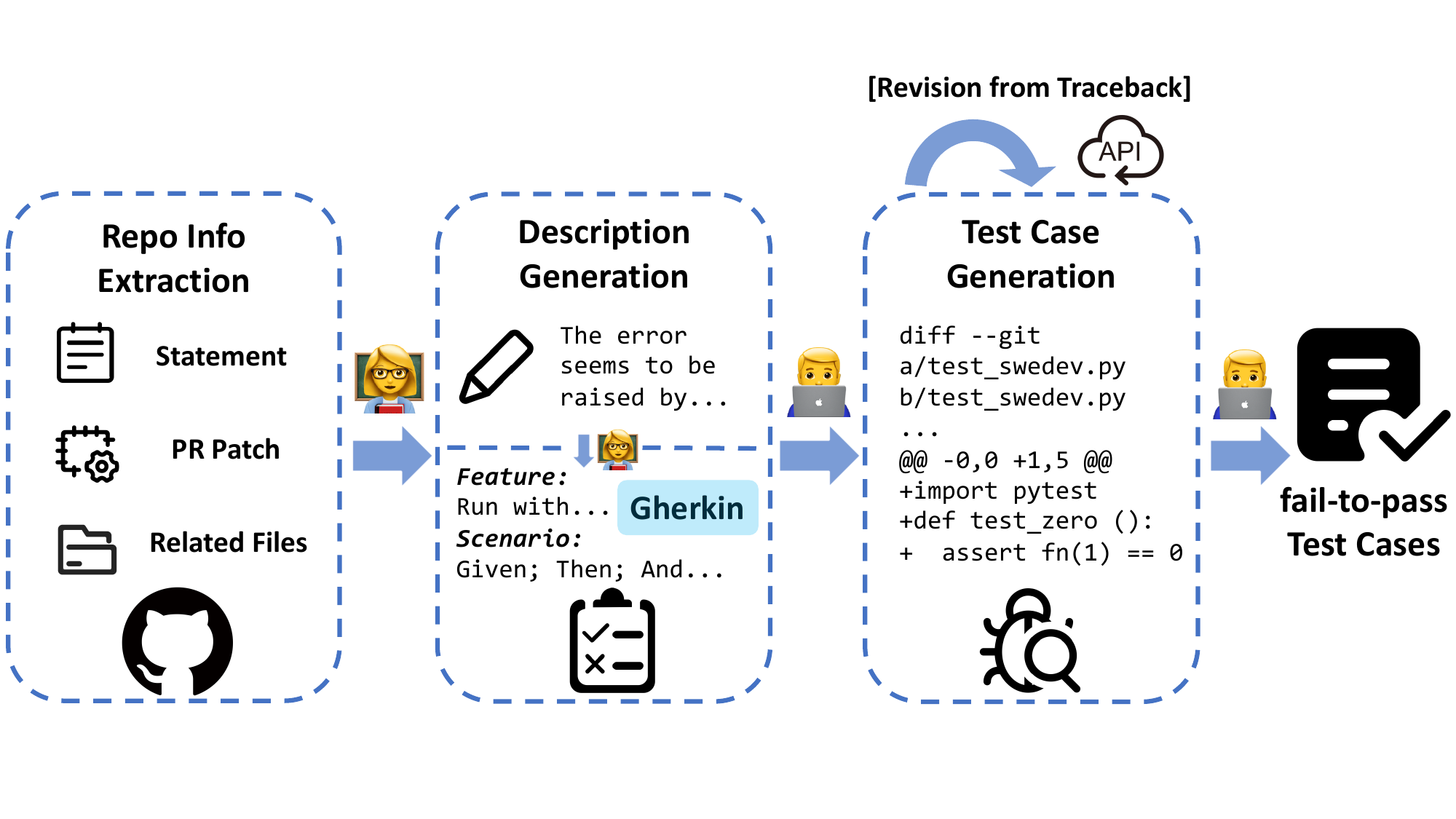}
  \caption{Pipeline for test case generation, divided into description generation and code generation phases. The pipeline begins with extracting repository information, followed by generating Gherkin scenarios and then detailed test cases. An optional revision step leverages traceback errors to refine the generated test cases. The final output includes fail-to-pass test cases.}
  \label{fig:testcase_flow}
\end{figure*}

\section{Related Works}

\paragraph{SWE Dataset Construction.} Prior benchmarks like SWE-bench~\citep{jimenez2024swebenchlanguagemodelsresolve}, DevEval~\citep{li2024devevalmanuallyannotatedcodegeneration}, EvoCodeBench~\citep{li2024evocodebenchevolvingcodegeneration} and Commit0~\citep{zhao2024commit0librarygenerationscratch} crawl from a large dataset and manually annotate the desired instance, which is labor consuming. There are works trying to filter existing fail-to-pass test cases like SWE-Gym~\citep{pan2024trainingsoftwareengineeringagents}, which annotates 2.4k instances by hand. Filtering out from a large dataset~\citep{golubev2024search} is effective but will leave out many useful issues only because the corresponding PR does not contain test cases. Therefore, automatically generating test cases from issue descriptions becomes essential.

\paragraph{SWE Agentic Framework.} SWE agent framework focuses on two mainstream types: Interaction-based frameworks and pipeline-based frameworks. Interaction-based frameworks like OpenHands~\citep{wang2024openhandsopenplatformai}, SWE-Agent~\citep{yang2024sweagentagentcomputerinterfacesenable}, Learn-By-Interact~\citep{su2025learnbyinteractdatacentricframeworkselfadaptive} and \citet{wang2024executablecodeactionselicit} usually pre-define a set of agent-computer interfaces (ACIs) to help model manipulate the environment. Meanwhile, pipeline-based frameworks will design the whole process into several steps, like Agentless~\citep{xia2024agentlessdemystifyingllmbasedsoftware}, CoreR~\citep{chen2024coderissueresolvingmultiagent}, MarsCode~\citep{liu2024marscodeagentainativeautomated} and SuperCoder2.0~\citep{gautam2024supercoder20technicalreportexploring}. The agent will generate patches by employing fault localization, patch generation and major voting. There are also techniques during inference time like MCTS~\citep{antoniades2024swesearchenhancingsoftwareagents} and critic-guided generation~\citep{badertdinov2024scaling} and model assembly~\citep{zhang2024diversityempowersintelligenceintegrating}. While pipeline-based process may benefit from specification, it cannot generalize to other coding tasks. On the contrary, interaction-based frameworks can do general tasks with natural language as instruction.

%% file: paras/method.tex
\section{SWE-Dev: Building Software Tasks at Scale}

In this part, we developed a systematic strategy to build software tasks at scale. The core of \textsc{SWE-Dev} is to collect repositories and task instances at scale and then develop a LLM-based automated test-case generation approach. 
This scalable approach enables us to construct comprehensive training datasets for SWE tasks and enhance performance through test case-driven trajectory sampling as Figure \ref{fig:testcase_flow} shows.
Based on the systematic pipeline for scalable dataset construction and trajectory optimization, we build SWE-Dev agents.

\subsection{Instance Collection}

We began by crawling metadata for 240k PyPI packages containing GitHub URLs, filtering for repositories with Stars $\geq$ 5 and PRs $\geq$ 3, resulting in a subset of 59k repositories. Due to network constraints, machine capacity and intricate dependency management, we successfully downloaded 10,416 repositories. Following the methodology outlined in \citet{jimenez2024swebenchlanguagemodelsresolve} with minor modifications, we extracted 88k instances in total.

\begin{figure}[t]
  \centering
  \includegraphics[width=\columnwidth]{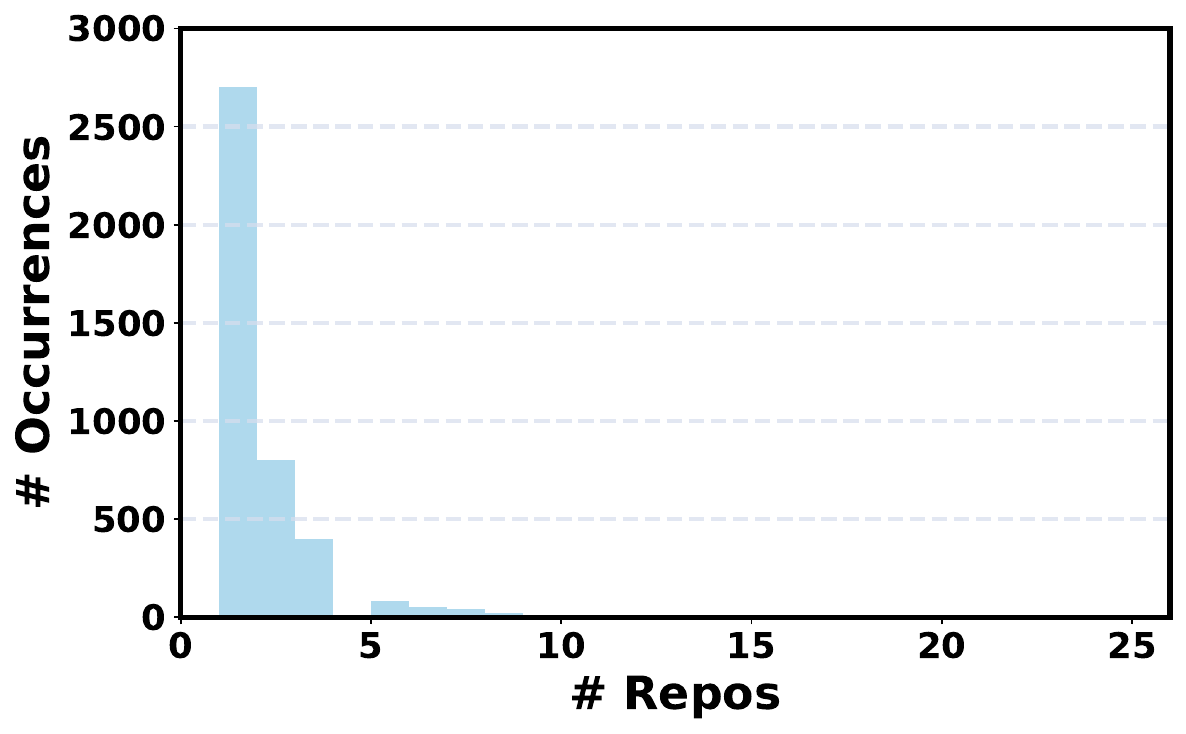}
  \caption{Distribution of instances per repository in the training dataset. The majority of repositories contribute fewer than five instances, highlighting the dataset's diversity across a wide range of repositories.}
  \label{fig:repo_distribution}
\end{figure}

To refine this dataset, we applied rule-based filtering to adjust patch lengths to fit model context windows and eliminated trivial or irrelevant issues while maintaining diversity. Ultimately, we retained 38k instances from 4,413 repositories as the training set. As shown in Figure~\ref{fig:repo_distribution}, over 4,000 repositories contain fewer than five instances, significantly enhancing the dataset’s diversity.

\subsection{Automatic Test Case Generation}

While it is relatively straightforward to crawl PRs containing golden patches from the Internet, these patches often lack corresponding test cases. This absence prevents the validation of trajectory correctness and makes reinforcement learning techniques infeasible. To address this, we leveraged LLMs to generate test cases with necessary context.

In detail, our test case generation pipeline consists of four steps. The process begins with \textbf{extracting relevant information} from the codebase, including contextual code snippets and metadata, which serve as the foundation for subsequent steps. Next, the extracted information is used for \textbf{generating Gherkin descriptions}, where the model produces structured, high-level scenarios. These descriptions adopt the Gherkin syntax, using special keywords to provide structure and meaning, making the scenarios concise and easily interpretable. Following this, the descriptions are utilized for the \textbf{generation of test cases}, which leverages the contextual details to create robust and meaningful test cases. After, the generated test case can undergo an optional \textbf{revision phase}, where they are reviewed and refined to ensure correctness.

This phased approach reflects the inherent structure of how LLMs generate test cases. The model extracts a scenario from the provided context and, using the code snippet as a reference, produces the corresponding test case. Merging the first three steps into a one often results in irrelevant or incoherent test cases. By introducing fine-grained instructions, we provide clearer guidance, significantly improving the accuracy and contextual relevance of the generated outputs.

Due to computing budget, we split part of the dataset to synthesize test cases. From a dataset of 26k instances, we generated 2,097 fail-to-pass functions. Additionally, we integrated existing test cases from our dataset, resulting in a total of 4,630 test cases. Notably, most unsuccessful instances in our approach are due to environmental constraints rather than fundamental limitations of the methodology, further underscoring the feasibility of generating test cases via our approach.

\subsection{Dataset Statistics}

To evaluate the quality of our constructed dataset, we conducted three experiments: (1) pass rate analysis between description and code synthesis (2) comparison of filtered issues with open-source datasets (3) validation of test cases. Together, these experiments highlight the strengths of our dataset in generating high-quality test cases and its utility for downstream tasks.

These experiments collectively highlight the strengths of our dataset in generating high-quality test cases and its utility for downstream tasks.

\paragraph{Pass Rate Analysis Between Description and Code Synthesis.}
As shown in Table~\ref{tab:model_statistics}, we evaluated several models and a mixed approach that combines Llama for description generation and Qwen-Coder (trained primarily on code data) for code synthesis. The mixed approach significantly outperformed individual models in generating fail-to-pass (F2P) functions and instances with test cases, highlighting the importance of leveraging domain-specific strengths for each task.

\begin{table}[htbp]
\centering
\begin{tabular}{l|r|r|r|r}
\hline
\textbf{Model} & \textbf{\#w/test} & \textbf{F2P} & \textbf{F2F} & \textbf{\#w/F2P}\\ 
\hline
Llama & 757 & 185 & 1273 & 73 \\ 
Mistral & 767 & 151 & 990 & 81 \\ 
Qwen & 793 & 190 & 1185 & 83 \\ 
Mix & 828 & 237 & 1408 & 92 \\ 
\hline
\end{tabular}
\caption{Model Comparison for Test Case Synthesis. \textit{\#w/test} and \textit{\#w/F2P} denote the number of instances with test cases and with F2P test cases respectively. Models include Llama-3.1-70B-Instruct, Qwen-2.5-Coder-32B-Instruct, and Mistral-Large-Instruct-2407. \textit{F2P} and \textit{F2F} represent the total test functions generated during synthesis. The \textit{Mix} model uses Llama for description generation and Qwen for code generation.}
\label{tab:model_statistics}
\end{table}

\paragraph{Comparison with Open-Source Datasets.}
We further validated the quality of our generated test cases by comparing models trained on issues from different datasets. As shown in Table~\ref{tab:random_sft_results}, the inclusion of our generated test cases achieves quality comparable to existing open-source dataset~\citep{badertdinov2024scaling}.

\begin{table}[ht]
\centering
\begin{tabular}{lc}
\hline
\textbf{Dataset} & \textbf{Resolve Rate} \\
\hline
Nebius/SWE-bench-extra & 15.0 \\
SWE-Gym/SWE-Gym & 13.8 \\
SWE-Dev (kept) & 15.4 \\
SWE-Dev (filtered) & 12.4 \\
\hline
\end{tabular}
\caption{Performance Comparison Across Different Training Datasets. All trajectories are unevaluated. \textit{SWE-Dev (kept)} refers to prompts retained after filtering, while \textit{SWE-Dev (filtered)} corresponds to discarded prompts. \textit{Random} means randomly selecting from trajectory pool.}
\label{tab:random_sft_results}
\end{table}

\paragraph{Validation of LLM-Generated Test Cases.}
We compared the effectiveness of our dataset with other widely-used datasets, such as Nebius/SWE-bench-extra. As shown in Table~\ref{tab:perform_dataset}, models trained on our dataset achieve comparable or slightly better performance than those trained on Nebius.

\begin{table}[ht]
\centering
\begin{tabular}{lccc}
\hline
\textbf{Model} & \textbf{900-RFT} & \textbf{500-RFT} \\
\hline
Nebius & 15.80 & 14.87 \\
SWE-Dev & 15.87 & 15.00  \\
Random & 14.67 & 13.87  \\
\hline
\end{tabular}
\caption{Performance of RFT across different datasets. We choose a 7B base model with 13.6\% resolve rate. \textit{k-RFT} denotes fine-tuning with $k$ correct trajectories. \textit{Random} means sampling randomly from our pool. Each configuration was independently run three times.}
\label{tab:perform_dataset}
\end{table}

Using 900 trajectories for rejection sampling fine-tuning, models trained on Nebius achieved a slightly higher mean accuracy of 15.80, while our dataset achieved 15.87. Even with 500 trajectories, our dataset remained competitive with a mean accuracy of 15.00 compared to Nebius’s 14.87. This demonstrates that our custom test cases are also effective in verifying whether trajectories comply with the requirement from issues.

\begin{table*}[h]
\centering
\small
\renewcommand{\arraystretch}{1.25}
\begin{tabular}{@{}lccc@{}}
\toprule
\textbf{Method}  & \textbf{Framework} & \textbf{Model} & \textbf{Resolve Rate} \\
\midrule
\rowcolor[gray]{0.9} 
\multicolumn{4}{c}{\textbf{Proprietary Models}} \\
OpenAI-GPT-4o~\cite{openai_gpt4o_systemcard_2024} & Pipeline  & GPT-4o & 33.2\% \\
 
OpenAI-o1-preview~\cite{openai_o1_systemcard_2024} & Pipeline  & OpenAI-o1-preview & 41.3\% \\
OpenAI-o1~\cite{openai_o1_systemcard_2024} & Pipeline & OpenAI-o1 & 48.9\% \\
Moatless Tools~\cite{orwall2024moatless} & Moatless Tools & claude-3-5-sonnet-20241022 & 39.0\% \\
AutoCodeRover~\cite{zhang2024autocoderoverautonomousprogramimprovement}  & AutoCodeRover & GPT-4o (2024-05-13) &  38.4\% \\
AutoCodeRover v2.0~\cite{zhang2024autocoderoverautonomousprogramimprovement}  & AutoCodeRover & claude-3-5-sonnet-20241022 &  46.2\% \\
Agentless-1.5~\cite{xia2024agentlessdemystifyingllmbasedsoftware}  & Agentless & GPT-4o (2024-05-13) & 38.8\% \\
Agentless-1.5~\cite{xia2024agentlessdemystifyingllmbasedsoftware}  & Agentless & claude-3-5-sonnet-20241022 & 50.8\% \\

OpenHands + CodeAct v2.1~\cite{wang2024executablecodeactionselicit}  & OpenHands & claude-3-5-sonnet-20241022 & \textbf{53.0}\% \\

\rowcolor[gray]{0.9} 
\multicolumn{4}{c}{\textbf{Open-source Models}} \\

SWE-Llama-13B~\cite{jimenez2024swebenchlanguagemodelsresolve}   & RAG & Code Llama-13B & 1.2\% \\

SWE-Gym-7B~\cite{pan2024trainingsoftwareengineeringagents}   & OpenHands & Qwen2.5-Coder-7B-Instruct & 10.6\% \\
\textbf{SWE-Dev-9B(Ours)} & OpenHands & GLM-4-9b-chat & 
$13.6\% (\uparrow 12.0\%)$\\
\textbf{SWE-Dev-8B(Ours)} & OpenHands & Llama-3.1-8B-Instruct & $18.0\% (\uparrow 16.8\%)$\\
SWE-SynInfer-7B~\cite{ma2024lingmaswegptopendevelopmentprocesscentric}  & Agentless & Qwen2.5-Coder-7B & 18.2\% \\
\textbf{SWE-Dev-7B(Ours)} & OpenHands & Qwen2.5-Coder-7B-Instruct & 
$\mathbf{23.4\% (\uparrow 21.6\%)}$\\

\midrule

SWE-Gym-32B~\cite{pan2024trainingsoftwareengineeringagents}  & OpenHands & Qwen2.5-Coder-32B-Instruct & 20.6\% \\

SWE-SynInfer-72B~\cite{ma2024lingmaswegptopendevelopmentprocesscentric}  & Agentless & Qwen2.5-72B-Instruct &
30.2\% \\
SWE-Fixer-72B*~\cite{xie2025swefixertrainingopensourcellms}  & Agentless & Qwen2.5-72B & 32.8\% \\

\textbf{SWE-Dev-32B(Ours)}  & OpenHands & Qwen2.5-Coder-32B-Instruct & $\mathbf{36.6\% (\uparrow 30\%)}$ \\

\bottomrule
\end{tabular}
\caption{Comparison of resolve rates on the SWE-bench-Verified dataset. The table categorizes models into baselines and SWE agents, showcasing their performance. SWE-Dev models attain top-tier results within the realm of open-source models and concurrently exhibit robust performance among closed-source models. The relative improvement ($\uparrow$) for our models is calculated with respect to their respective base models.}
\label{tab:resolve_rate}
\end{table*}

By integrating fine-grained description generation, code generation, and execution-based validation, we ensure that the generated test cases are both accurate and contextually relevant. This approach not only bridges the gap of missing test cases in PR data but also significantly enhances the dataset’s utility for downstream training tasks.

%% file: paras/experiments.tex
\section{Experiments}

In this section, we will introduce our experiments on SWE agent tuning. As Table~\ref{tab:resolve_rate} shows, our models achieve state-of-the-art results among open-source models, with SWE-Dev-32B achieving a 36.6\% resolve rate, increasing 30\% resolve rate compared to our base model. Furthermore, \textsc{SWE-Dev} demonstrate competitive performance against closed-source models, narrowing the gap with leading systems such as OpenHands + CodeAct v2.1. These results highlight the effectiveness of our pipeline and training strategies.

We focus here on reporting key results from standard fine-tuning and reinforcement learning workflows. The impact of training and inference scaling will be discussed in the following section.

\subsection{Experiment Setup}

\paragraph{Dataset and Agentic Scaffold.}
We utilize the DeepSeek-V3~\cite{deepseekai2024deepseekv3technicalreport} for trajectory generation from both the Nebius and SWE-Dev datasets, and we take OpenHands, a ReAct-like~\citep{yao2023reactsynergizingreasoningacting} framework. After sampling, we obtained 17k unevaluated trajectories and 2.3k correct trajectories. Specifically, 914 of the correct trajectories originate from Nebius, while 78\% of the unevaluated trajectories are derived from \textsc{SWE-Dev}. For the OpenHands configuration, we set \texttt{iteration=30} and \texttt{max\_tokens=32k}. During offline RL training, we incorporate trajectories from DeepSeek-V3 and on-policy models.

\paragraph{Training Configuration.}  
We adopt the Qwen-2.5-Coder~\citep{hui2024qwen25codertechnicalreport}, Llama-3.1~\citep{grattafiori2024Llama3herdmodels} and GLM-4~\citep{glm2024chatglmfamilylargelanguage} series as our base models and leverage the OpenRLHF~\citep{hu2024openrlhfeasytousescalablehighperformance} framework for training. For datasets with $\leq$ 10k samples, we use a learning rate of $1\text{e-}5$ and a batch size of 32; for larger datasets, we maintain the same learning rate but increase the batch size to 64. The training process spans 4 epochs for the 7B model and 8 epochs for the 32B model. To enable long-context training, we apply ring-attention~\citep{liu2023ringattentionblockwisetransformers} with 8 heads.

\paragraph{Evaluation Metrics}  
For evaluation, we employ SWE-bench-Verified, a human-validated subset of 500 instances curated by OpenAI. This benchmark is carefully selected from SWE-bench~\citep{jimenez2024swebenchlanguagemodelsresolve}, ensuring high-quality evaluation. Model-generated code patches are assessed using developer-written unit tests, with accuracy defined as the percentage of successfully resolved instances. For all experiments, we set \texttt{iteration=30} and \texttt{max\_tokens=32k}, except for inference scaling experiments, where \texttt{max\_tokens=160k} is used.

\subsection{Main Results}
Table~\ref{tab:resolve_rate} presents the comparison of resolve rates on the SWE-bench-Verified benchmark. We evaluate two major approaches: methods leveraging fine-tuning of open-source models and the direct utilization of proprietary models.

Compared to existing methods based on fine-tuning open-source models, \textsc{SWE-Dev} achieves the highest performance without employing any verifiers or sampling strategies. SWE-Dev-32B outperforms approaches SWE-Syninfer-72B~\cite{ma2024lingmaswegptopendevelopmentprocesscentric} and SWE-Fixer-72B~\cite{xie2025swefixertrainingopensourcellms} which have a resolve rate of 36.6\%, achieving a 6.4\% improvement while using fewer parameters. With comparable model parameters, \textsc{SWE-Dev} significantly outperforms SWE-Gym~\cite{pan2024trainingsoftwareengineeringagents} and SWE-Syninfer by over 12.8\% and 5.2\%.

Furthermore, by fine-tuning an open-source model that is substantially inferior to proprietary models, our method achieves an impressive 3.4\% performance improvement over GPT-4o~\citep{openai2025competitiveprogramminglargereasoning} on the SWE-bench-Verified benchmark. Moreover, this approach demonstrates significant progress in effectively addressing complex SWE tasks with open-source models.

\input{paras/rs_rl}

%% file: paras/rs_rl.tex
\subsection{Agentic Post-Training Optimization}

\begin{figure}[htbp]
  \centering
\includegraphics[width=0.8\columnwidth]{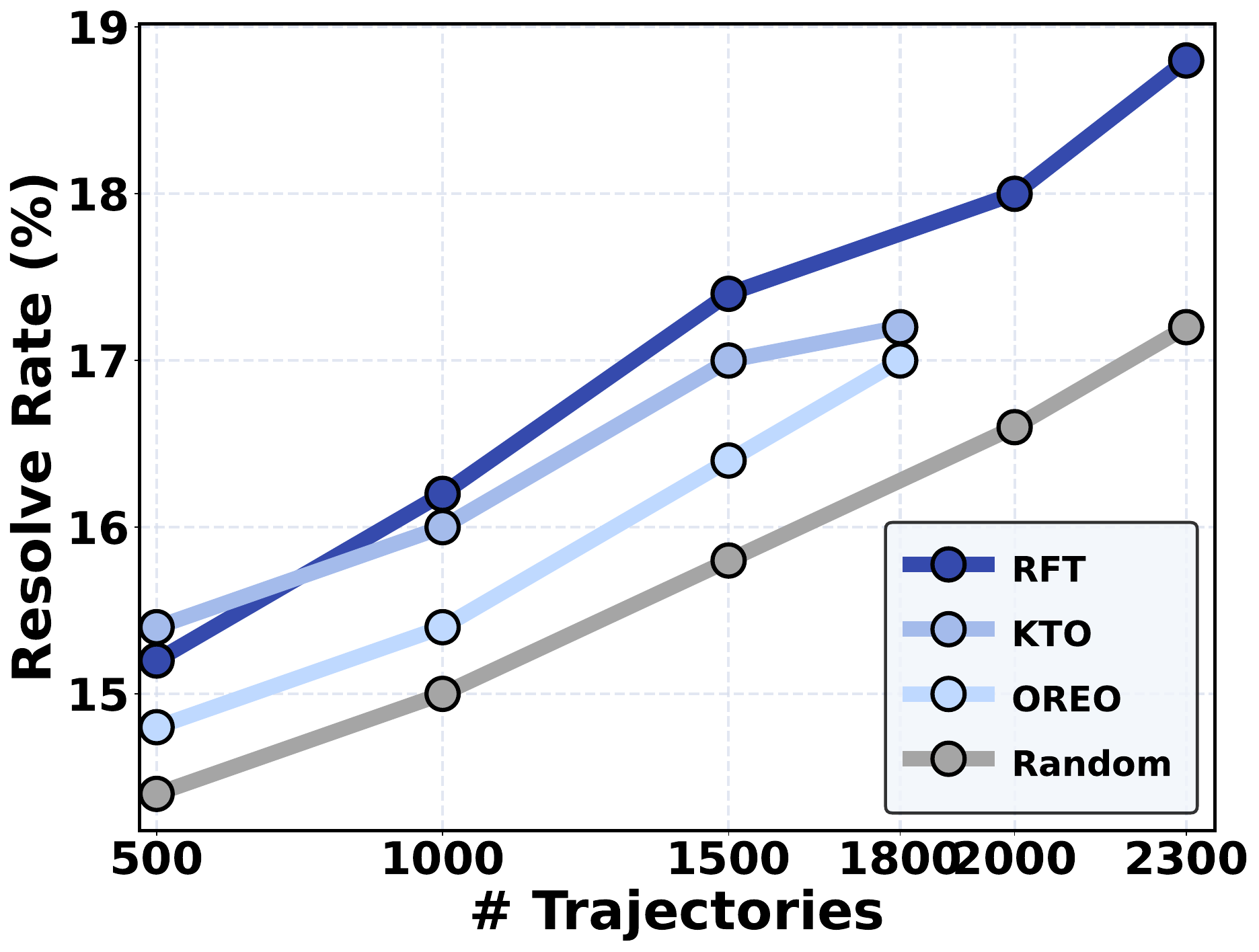}
  \caption{Model performance across different training methods. \textit{Random} denotes supervised fine-tuning with pass@1 trajectories. Other experimental settings remain consistent with those described above.}
  \label{fig:rs_scaling}
\end{figure}

\paragraph{Comparison of Training Methods.}
To further enhance model performance, we leveraged instances with fail-to-pass test cases to investigate rejection sampling fine-tuning as \citet{chen2023fireactlanguageagentfinetuning} and \citet{zeng2023agenttuningenablinggeneralizedagent}, with two offline RL methods: KTO and OREO. Our findings, as illustrated in Figure~\ref{fig:rs_scaling}, demonstrate the significant advantages and limitations of these techniques.

With only 1,000 trajectories, rejection sampling achieves a 16.2\% resolve rate, surpassing the performance of a larger-scale unfiltered dataset. As the number of trajectories increases, rejection sampling further narrows the performance gap, reaching an 18.0\% resolve rate with 2,000 trajectories. This approach is valuable in scenarios where training data is costly or computational resources are limited.

We also evaluated two offline RL methods: OREO and KTO. For both methods, positive examples are derived from DeepSeek-V3, and negative examples are sampled from a mixture of DeepSeek-V3 and on-policy outputs. Despite their theoretical potential, these offline methods demonstrated limited performance improvements compared to RFT.

OREO showed small but consistent gains over random sampling, achieving a 17.0\% resolve rate with 1,800 trajectories. However, its performance remained below that of RFT. KTO performed slightly better than OREO, achieving a 17.2\% resolve rate with 1,800 trajectories. While KTO marginally outperformed OREO, it still lagged behind RFT in overall effectiveness.

The relatively modest improvements observed with offline RL methods suggest that their reliance on on-policy negative examples and off-policy positive examples may limit their efficacy. On-policy negatives could introduce noise, while the positive examples might lack sufficient diversity to drive meaningful policy optimization.

\paragraph{Hybrid Training with RFT+OREO.}

To further explore the potential of combining different training strategies, we conducted hybrid training experiments by integrating RFT and OREO. Specifically, we divided the 2.3k deduplicated correct trajectories into two complementary subsets: one portion was allocated to RFT, while the remaining data was utilized for OREO, ensuring an efficient use of the available training data.

Importantly, in each run, the RFT+OREO hybrid approach utilized the entire training set, ensuring that the total amount of data seen by the hybrid model matched the full dataset. For comparison, the Random baseline was trained with the same number of pass@1 trajectories as RFT, while other experimental settings were kept consistent. The results are presented in Figure~\ref{fig:hybrid_rs_scaling}.

The RFT-only approach consistently outperformed all other configurations, achieving the highest resolve rate across all subsets. With 2300 RFT trajectories, the model achieved a resolve rate of 21.2\%, surpassing all hybrid and OREO-only configurations. This highlights RFT's strong ability to utilize high-quality data to maximize performance.

%% file: paras/scaling.tex
\section{Scaling SWE Agents via Training and Inference Expansion}

We explore how SWE agent performance scales along two critical dimensions: the amount of training data and the compute budget. Our findings reveal consistent trends aligned with established scaling laws, and shed light on the cost–performance trade-offs in agent-based settings.

\begin{figure}[t]
\includegraphics[width=1.0\columnwidth]{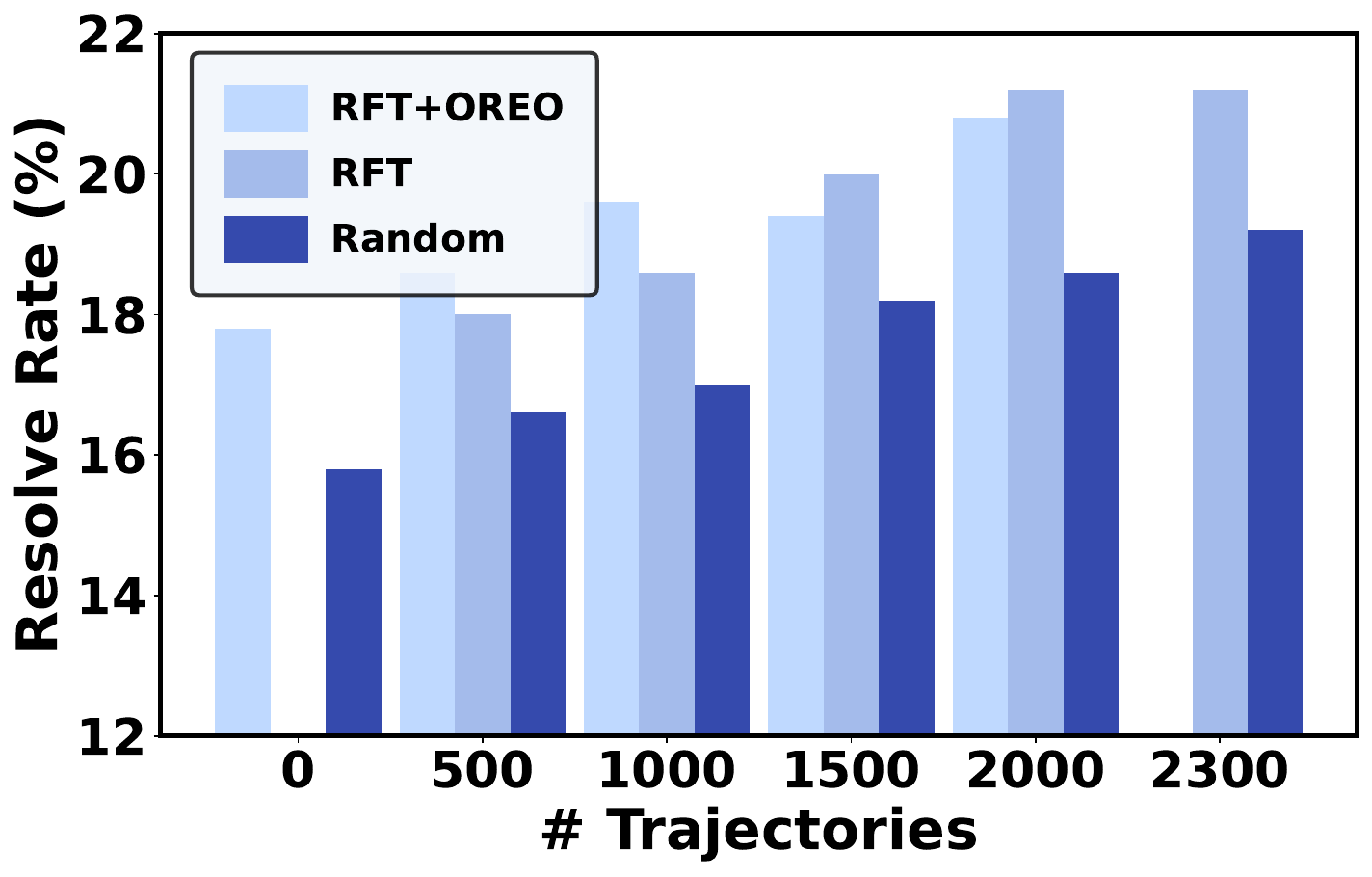}
  \caption{Model performance with hybrid training. In each run, RFT is applied to a subset of the data (measured in trajectories), followed by OREO training on the remaining data. The Random baseline uses the same number of pass@1 trajectories as RFT.}
  \label{fig:hybrid_rs_scaling}
  \vspace{-1.5em}
\end{figure}

\input{paras/data_scaling}

\input{paras/inference_scaling}

%% file: paras/data_scaling.tex
\subsection{Training Data Scaling}

\begin{figure*}[htbp]
  \centering
  \begin{subfigure}[t]{0.38\textwidth}     \centering
    \includegraphics[width=\textwidth]{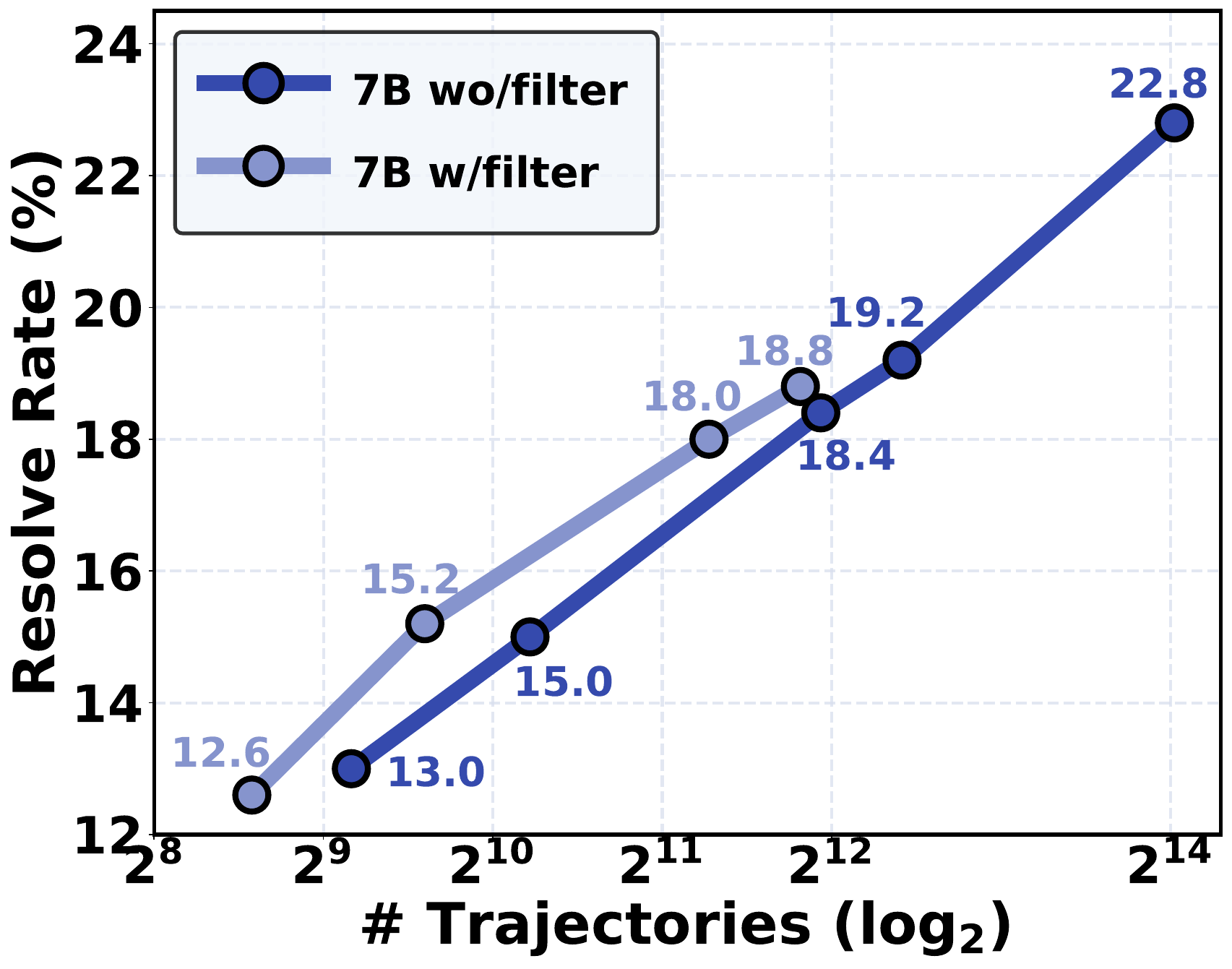}
    \label{fig:data_scaling_7b}
  \end{subfigure}
  \hspace{2em}
  \begin{subfigure}[t]{0.38\textwidth}     \centering
    \includegraphics[width=\textwidth]{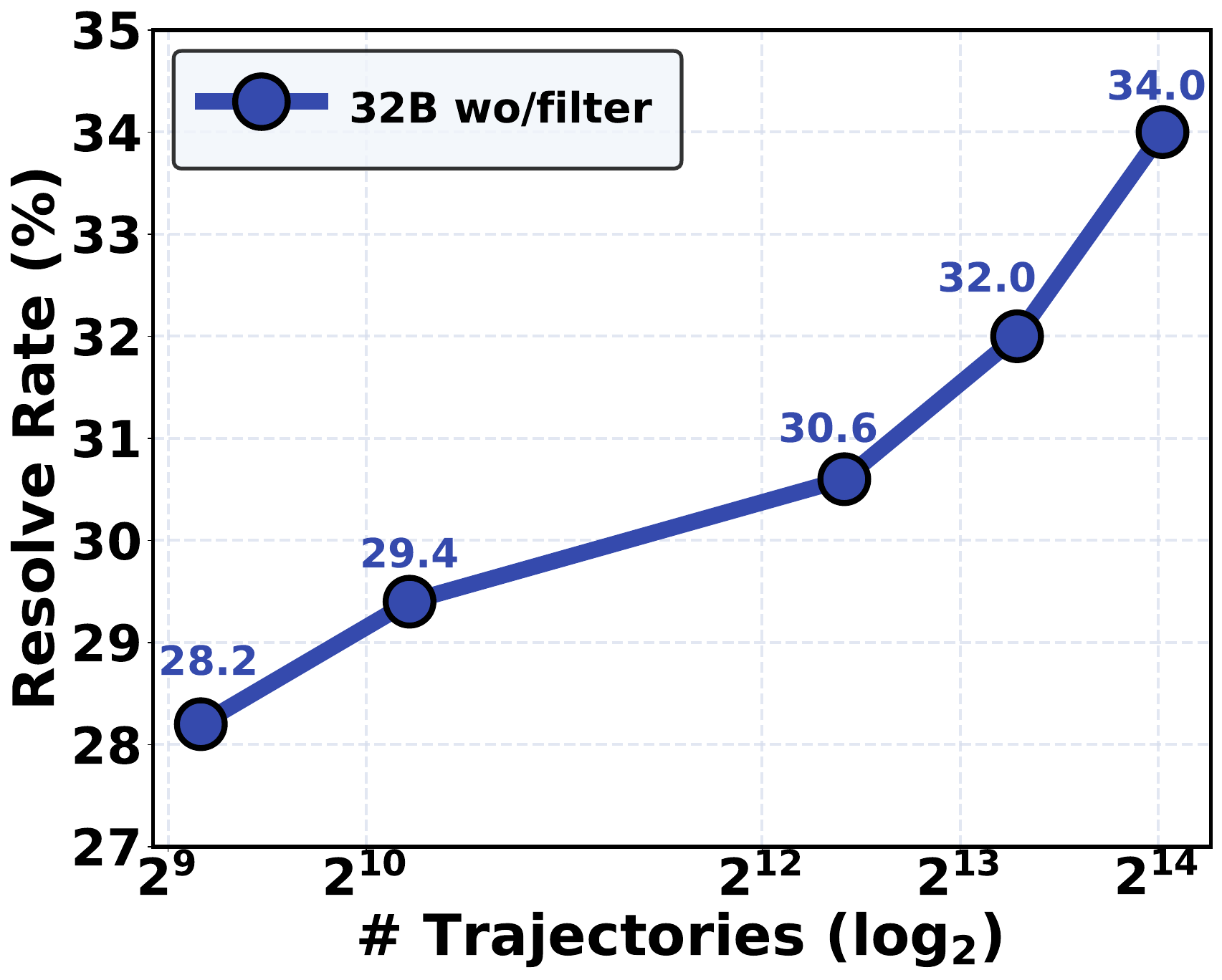}
    \label{fig:data_scaling_14b}
  \end{subfigure}
  \vspace{-1em}
  \caption{Resolve rates across different training data sizes. Both 7B and 32B models exhibit performance improvements with increased training trajectories.}
  \label{fig:data_scaling}
\end{figure*}

\begin{figure*}[!h]
  \centering
  \begin{subfigure}[t]{0.40\textwidth}     \centering
    \includegraphics[width=\textwidth]{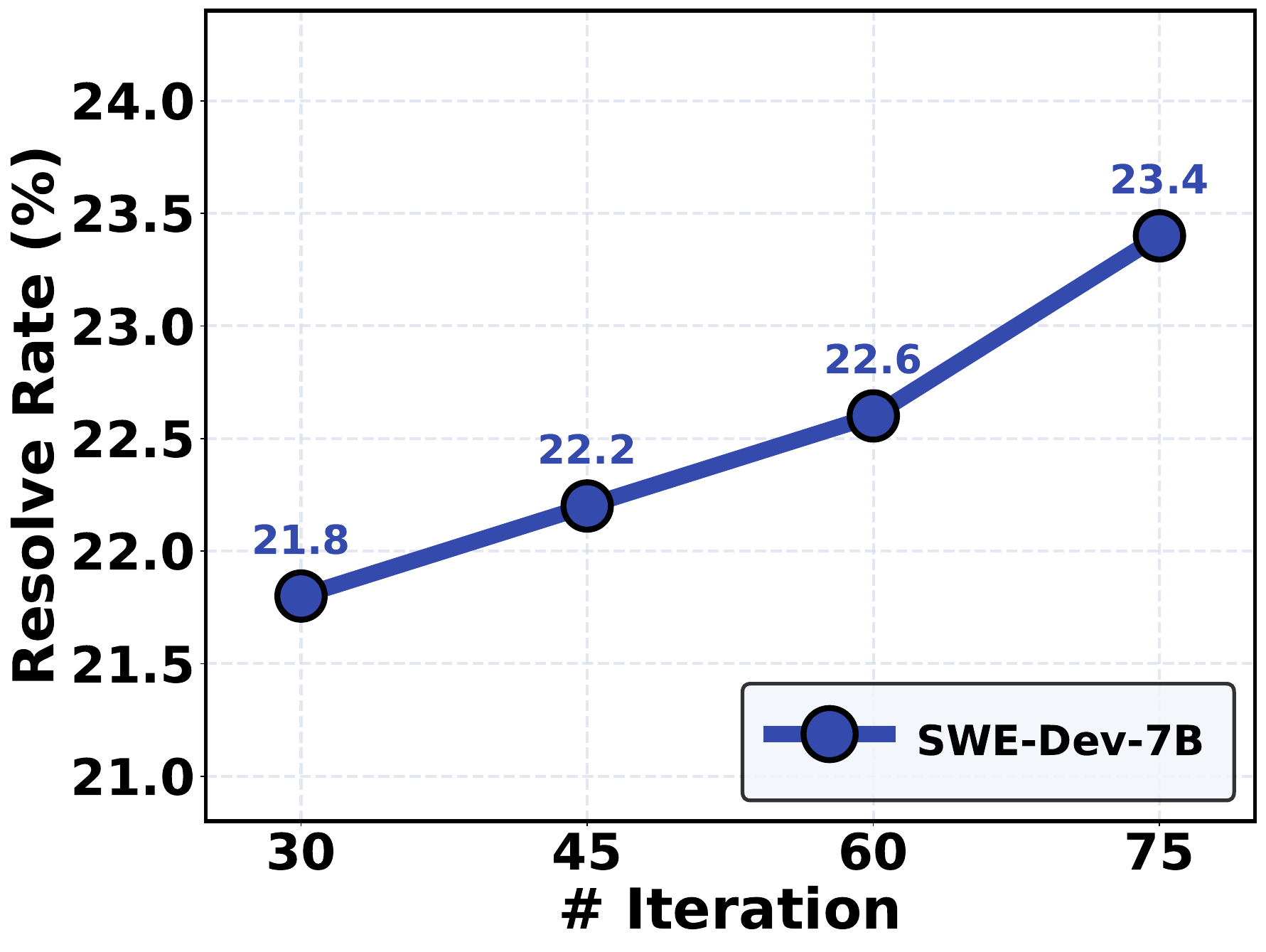}
    \label{fig:infer_swe_7b}
  \end{subfigure}
  \hspace{2em}
  \begin{subfigure}[t]{0.38\textwidth}     \centering
    \includegraphics[width=\textwidth]{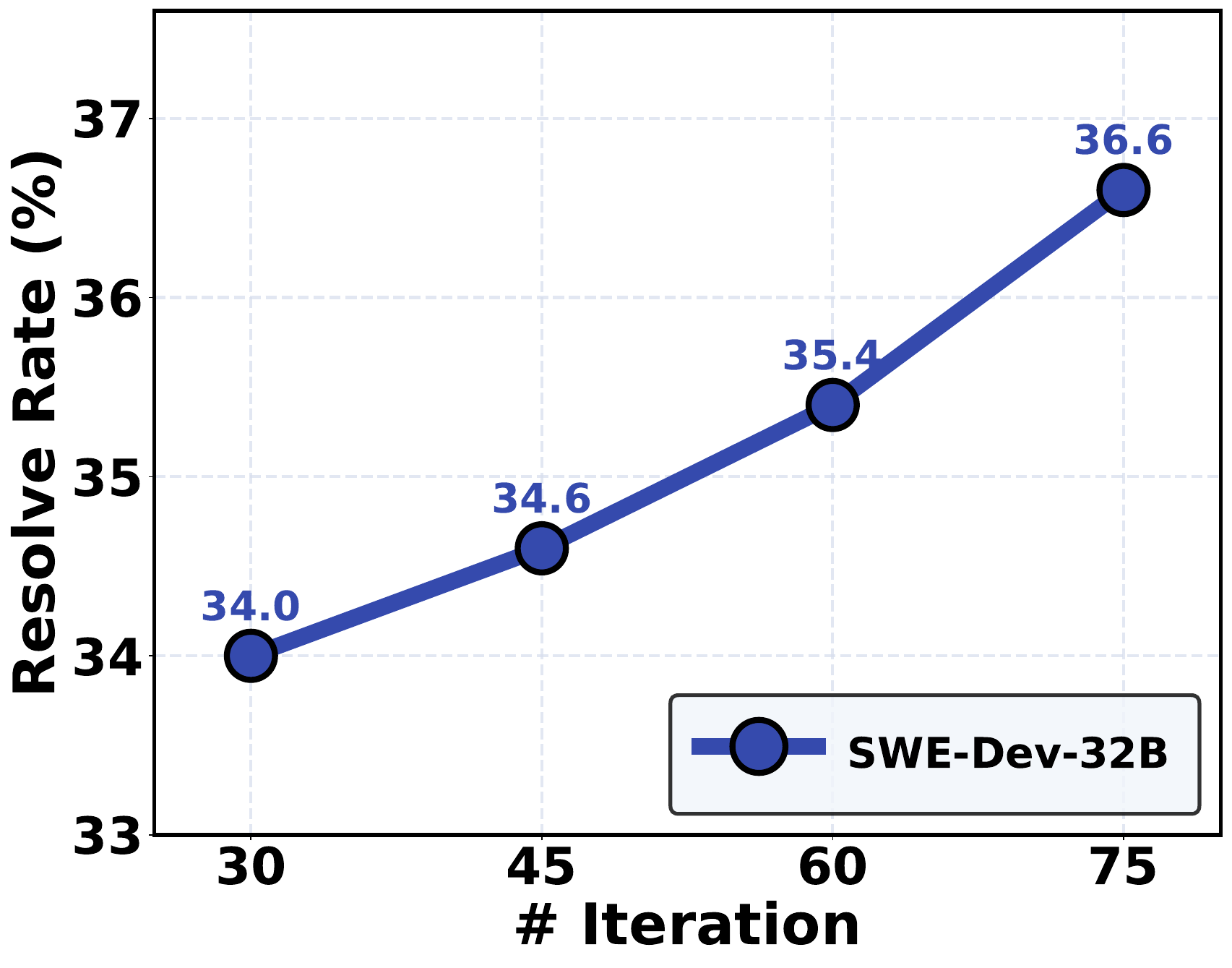}
    \label{fig:infer_swe_32b}
  \end{subfigure}
  \vspace{-1em}
  \caption{Resolve rates across different numbers of interaction rounds (30–75). Larger models benefit more significantly from extended iteration scaling, though diminishing returns are observed at higher rounds.}
  \label{fig:inference_scaling}
\end{figure*}

To investigate the impact of data size on model performance, we conducted a scaling analysis by progressively increasing the number of training trajectories. Notably, we filtered trajectories that did not terminate at the given iterations, ensuring that the remaining trajectories were unevaluated. As illustrated in Figure~\ref{fig:data_scaling}, the model's accuracy increases steadily with the logarithm of the data size, following a near-linear trend in log-log space. For instance, the 7B model achieved 13.0\% accuracy with only 574 trajectories. When the data size was scaled up to 16,639 trajectories, the accuracy improved significantly to 22.8\%.

This trend underscores a strong correlation between the quantity of training data and the model's generalizable ability. The observed linearity in the curve suggests that the model efficiently leverages additional data without exhibiting saturation within the explored range. These findings align with established scaling laws reported in \citet{kaplan2020scalinglawsneurallanguage} and \citet{hou2024does}, where performance predictably improves with increased data size, particularly in the low-data regime.

Moreover, we observe that the benefits of data scaling are more pronounced for larger models. The 32B model shows a sharper improvement compared to the 7B model, suggesting that larger capacity enables more efficient utilization of the available data. This observation echoes prior findings in model scaling literature, where overparameterized models can extract richer inductive biases from limited supervision. 

We further applied a filtering strategy to the training trajectories to enhance data quality. Specifically, we evaluated the model-generated patches against reference patches using an LLM-based comparison (Llama 3.1-70B-Instruct). The outputs were categorized into four groups: \textit{identical}, \textit{mostly identical}, \textit{partially identical}, and \textit{different}. Only trajectories labeled as \textit{identical} or \textit{mostly identical} were retained, preserving approximately 65\% of the original trajectories.

As shown in Figure~\ref{fig:data_scaling}, the filtered dataset achieves nearly identical performance to the unfiltered dataset, despite a significant reduction in training data size. This demonstrates that removing low-quality or irrelevant training samples like stuck in environment setup has minimal impact on model performance, as the retained data is of sufficiently high quality to offset the reduction in quantity.

%% file: paras/inference_scaling.tex
\subsection{Inference Time Scaling}


In addition to training data scaling, inference-time decisions also play a crucial role in overall model capability. In agent-based tasks, determining an appropriate compute budget is crucial for balancing task success and resource efficiency~\citep{brown2024largelanguagemonkeysscaling}. Existing works also demonstrate that increasing test-time compute leads to better performance in math reasoning tasks~\cite{openaio1,hou2025advancing}. Traditional evaluation metrics, such as pass@k, typically focus on multiple independent attempts, which may cost to a large computing budget for SWE evaluation. To address this, we investigate the impact of iteration scaling—progressively increasing the number of interaction rounds within a single run. This approach provides a more natural and efficient alignment with the iterative reasoning inherent in these tasks.

In this experiment, we directly change the RoPE embedding from 32k to 160k and observed that the resolve rate of the 7B model dropped from 22.8\% to 21.8\%, while the 32B model showed no significant decline. The results, presented in Figure~\ref{fig:inference_scaling}, demonstrate how resolve rate improves as the number of interaction rounds increases. We compare two configurations: SWE-Dev-32B, and SWE-Dev-7B. Our key findings include: 

\paragraph{Consistent Improvement with Iteration Scaling.}
For all models, increasing the number of interaction rounds leads to higher resolve rate. For instance, SWE-Dev-32B achieves 34.0\% at 30 rounds and 36.6\% at 75 rounds. This demonstrates that additional iterations allow the models to refine their reasoning and correct prior errors, effectively leveraging the iterative interaction process.

\paragraph{Diminishing Returns Beyond a Threshold.} 
Although the resolve rate increases with additional interaction rounds, the rate of improvement diminishes over time. For instance, the improvement observed between 30 and 45 rounds is significantly more pronounced than that between 45 and 75 rounds across all models. This suggests a practical upper limit to the benefits of iteration scaling, beyond which additional rounds yield minimal performance gains. These diminishing returns highlight the importance of calibrating iteration limits to balance computational costs with improvements.

%% file: paras/conclusion.tex
\section{Conclusion}
In this work, we introduce \textsc{SWE-Dev}, an open-source SWE agent equipped with a robust test case construction pipeline. We further investigate the impact of techniques such as rejection sampling and offline RL methods, while proposing iteration scaling as a novel approach to enhance inference-time performance. Our models are evaluated on the SWE-bench-Verified benchmark, achieving state-of-the-art results among open-source agents, with performance scores of 23.4\% for the 7B model and 36.6\% for the 32B model.

Our work provides a strong foundation for advancing SWE-focused LLM research by introducing an open-source model, and a comprehensive data generation pipeline. We hope these contributions inspire further innovation in developing robust, scalable, and efficient SWE agents capable of addressing real-world SWE challenges.

%% file: paras/limitation.tex
\section{Limitation}
While offline RL have shown promise, there is significant room for optimization. Incorporating online RL approaches, such as ArCHer~\citep{zhou2024archertraininglanguagemodel}, DigiRL~\citep{bai2024digirltraininginthewilddevicecontrol}, and WebRL~\citep{qi2025webrltrainingllmweb}, could further enhance performance by leveraging interaction-based frameworks and dynamic task environments~\citep{snell2023offlinerlnaturallanguage}. Future research could explore adaptive iteration strategies, where the number of rounds is dynamically adjusted based on task complexity or intermediate progress. Hybrid approaches that combine iteration scaling with other optimization techniques may also improve performance while maintaining computational efficiency.

%% file: paras/appendix.tex
\appendix
\onecolumn
\section{Appendix}
\label{sec:appendix}
\subsection{Prompt Templates}
\begin{tcolorbox}[
    colback=white, 
    colframe=prompt-color, 
    coltitle=black, 
    title=\textbf{Prompt for Patch Filtering}, 
    fonttitle=\bfseries, 
    arc=2mm, 
    fontupper=\footnotesize, 
    breakable, 
    enhanced, 
    coltitle=black 
]

You are a professional programmer. Given two git patches that aim to fix the same bug, your task is to analyze if the second patch correctly fixes the bug and identify any issues in it. Please ignore any changes to \texttt{reproduce\_error.py} files as they are only used for debugging purposes.

\textbf{Information provided:}
\begin{itemize}
    \item \textbf{Patch 1 (correct solution)}: \{patch1\}
    \item \textbf{Patch 2 (patch to evaluate)}: \{patch2\}
\end{itemize}

\textbf{Guidelines:}
\begin{enumerate}
    \item When comparing patches, focus only on code changes that affect actual functionality. Differences in logging or documentation can be ignored.
    \item The evaluation levels are:
    \begin{itemize}
        \item \textbf{identical}: Patch 2's functionality is exactly the same as Patch 1.
        \item \textbf{mostly}: Patch 2 implements most of the functionality correctly with minor differences.
        \item \textbf{partially}: Patch 2 only implements some of the required functionality.
        \item \textbf{different}: Patch 2's functionality is completely different or incorrect.
    \end{itemize}
    \item If Patch 2 is rated as \textbf{partially} or \textbf{mostly}, please specify the functional differences in the explanation.
\end{enumerate}

\textbf{Final Instructions:}
\begin{itemize}
    \item Provide your analysis and judgment in the following format:
\begin{tcolorbox}[colback=gray!5, colframe=gray, breakable]
[Explanation]

Explanation of the reason why the patch is judged as 'identical', 'mostly', 'partially', or 'different'.

[Judgment]

The judgment of the patch is 'identical', 'mostly', 'partially', or 'different'.
\end{tcolorbox}
\end{itemize}

\end{tcolorbox}


\begin{tcolorbox}[
    colback=white, 
    colframe=prompt-color, 
    coltitle=black, 
    title=\textbf{Prompt for Generating Naive NL Test Case Description}, 
    fonttitle=\bfseries, 
    arc=2mm, 
    fontupper=\footnotesize, 
    breakable, 
    enhanced, 
    coltitle=black 
]

You are a skilled test engineer. Your mission is to create a minimal, edge-case test scenario that rigorously validates the effectiveness of the patch. This test case must satisfy the following conditions:

\begin{enumerate}
    \item \textbf{Fail with the unpatched code}: Demonstrate the specific bug, issue, or limitation that the patch is designed to address. Ensure the test triggers this behavior reliably and consistently.
    \item \textbf{Pass with the patched code}: Confirm that the patch resolves the issue without introducing new problems or regressions.
\end{enumerate}

\textbf{Focus Areas:}
\begin{itemize}
    \item Exercise uncommon or edge-case code paths.
    \item Test for boundary conditions or unexpected input.
    \item Mimic realistic usage scenarios where the original behavior fails.
\end{itemize}

\textbf{Information provided:}
\begin{itemize}
    \item \textbf{Repository name}: \{\}
    \item \textbf{GitHub issue description}: \{\}
    \item \textbf{Correction patch}: \{\}
    \item \textbf{Hints Text}: \{\}
\end{itemize}

\textbf{Your Task:}
\begin{itemize}
    \item Briefly analyze the problem description and hints text to identify \textbf{where the issue lies and what should be fixed}.
    \item Write a concise test case description that reflects the modification introduced by the patch.
    \item Ensure the description targets the root cause of the issue and guides the generation of a challenging, edge-case test scenario.
\end{itemize}

\textbf{Important Notes:}
- Avoid unrelated information or greetings.
- Focus exclusively on the test case description and the problem analysis.

\end{tcolorbox}

\begin{tcolorbox}[
    colback=white, 
    colframe=prompt-color, 
    coltitle=black, 
    title=\textbf{Prompt for Generating Gherkin Description}, 
    fonttitle=\bfseries, 
    arc=2mm, 
    fontupper=\footnotesize, 
    breakable, 
    enhanced, 
    coltitle=black 
]

You are an experienced test engineer. Your task is to write a test following the Gherkin syntax based on the information provided below. This test must verify whether the correction patch in the repository correctly resolves the described issue.

\textbf{Key Points:}
- The test should \textbf{fail} with the unpatched code and \textbf{pass} with the patched code.

\textbf{Information Provided:}
\begin{itemize}
    \item \textbf{Repository name}: \{\}
    \item \textbf{GitHub issue description}: \{\}
    \item \textbf{Correction patch}: \{\}
    \item \textbf{Hints Text}: \{\}
    \item \textbf{Analysis for the test cases (for reference)}: \{\}
\end{itemize}

\textbf{Requirements:}
\begin{enumerate}
    \item Use the \textbf{Given-When-Then} structure of Gherkin syntax.
    \item Clearly describe:
    \begin{itemize}
        \item Preconditions (\texttt{Given}).
        \item Triggering events (\texttt{When}).
        \item Expected outcomes (\texttt{Then}).
    \end{itemize}
    \item Ensure the test logic is clear, concise, and covers all relevant scenarios.
    \item Avoid including unimportant test cases, such as modifications to README files.
\end{enumerate}

\textbf{Instructions:}
- Wrap each Gherkin test description in triple backticks (\texttt{```gherkin}).
- Example format:
\begin{verbatim}
```gherkin
{{YOUR DESCRIPTION}}
```
\end{verbatim}
\end{tcolorbox}

\begin{tcolorbox}[
    colback=white, 
    colframe=prompt-color, 
    coltitle=black, 
    title=\textbf{Prompt for Test Case Generation}, 
    fonttitle=\bfseries, 
    arc=2mm, 
    fontupper=\footnotesize, 
    breakable, 
    enhanced, 
    coltitle=black 
]

You are a test engineer. Given a GitHub issue description and the golden patch, your task is to build test cases that \textbf{reproduce the error} according to the patch. In detail, the test cases should reproduce the error in the issue description.

Your test case will run at the \textbf{root} of the project. Please be careful of the relative path to avoid path-related errors.

\textbf{Information provided:}
\begin{itemize}
    \item \textbf{Repository name}: \{\}
    \item \textbf{GitHub issue description}: \{\}
    \item \textbf{Hints Text}: \{\}
    \item \textbf{Correction patch}: \{\}
    \item \textbf{Project tree (file depth less than 3)}: \{\}
    \item \textbf{Test Case Description}: \{\}
    \item \textbf{Relevant code segments in the original version}: \{\}
\end{itemize}

\textbf{Steps to follow:}
\begin{enumerate}
    \item \textbf{Identify the incorrect code}: Analyze the provided information to locate the error that the patch addresses. Identify the required packages and the types of test cases to write.
    \item \textbf{Generate the test case}: Write test cases that will \textbf{fail without the correction patch} and \textbf{pass with the correction patch}. Each test case must be enclosed within \lstinline|<testcase></testcase>| tags. 
    \item Ensure that no additional execution beyond your test case is performed. Avoid unsafe commands or unnecessary changes to the project.
\end{enumerate}

\textbf{Format Requirements:}
\begin{itemize}
    \item Test Case:
    \begin{itemize}
        \item Wrap each test case in \lstinline|<testcase></testcase>| tags.
        \item Use triple backticks (\lstinline|```|) to enclose the test code within the \lstinline|<testcase></testcase>| tags.
        \item The test cases must be ready to run with \texttt{pytest} and should include any necessary mock data or fixtures.
    \end{itemize}
\end{itemize}

\textbf{Environment Information:}
\begin{itemize}
    \item \textbf{Python Version}: 3.9
    \item \textbf{Platform}: Ubuntu 22.04.5 LTS
    \item \textbf{Execution Command}: 
    \begin{lstlisting}[language=bash, breaklines=true]
python -m pytest --no-header -rA -p no:cacheprovider -W ignore::DeprecationWarning --continue-on-collection-errors --tb=short
    \end{lstlisting}
    \item \textbf{Execution Path}: Root directory of the project
\end{itemize}

\textbf{Example Solution:}

In \texttt{src/utils/csv\_utils.py}:
\begin{lstlisting}[language=Python, breaklines=true]
from CSVconverter.src.utils import csv
def read_csv_and_sum(filename):
    """Calculate the sum of all numbers in a CSV file"""
    total = 0
    with open(filename, 'r') as file:
        reader = csv.reader(file)
        for row in reader:
            total += row[0]
    return total
\end{lstlisting}

The code directly adds \texttt{row[0]} to \texttt{total} without validating if \texttt{row[0]} is an integer. If the CSV file contains non-numeric values (e.g., strings or empty fields), it will raise runtime errors like \texttt{TypeError} or \texttt{ValueError}. These errors match the problem statement. So I'll write test cases here.

\textbf{Fix Explanation:}
\begin{enumerate}
    \item It tries to convert \texttt{row[0]} to an integer using \texttt{int()}.
    \item If \texttt{row[0]} is not a valid integer, it skips that row using a \texttt{try...except} block.
\end{enumerate}

The goal is to write test cases that:
\begin{enumerate}
    \item Test case with non-numeric data in the CSV (should raise an error in the original code).
    \item Same test case should now correctly handle non-numeric rows and calculate the sum of valid numeric values.
\end{enumerate}

\textbf{Example Test Case:}
\begin{lstlisting}[language=Python, breaklines=true]
<testcase>
```python
import os
import pytest
from src.utils.csv_utils import read_csv_and_sum

@pytest.fixture
def create_csv_file():
    """Fixture to create a temporary CSV file for testing."""
    def _create_file(contents, filename="test.csv"):
        with open(filename, 'w') as f:
            f.write(contents)
        return filename
    yield _create_file
    # Cleanup after test
    if os.path.exists("test.csv"):
        os.remove("test.csv")

def test_valid_csv(create_csv_file):
    """Test case with valid numeric data."""
    filename = create_csv_file("1\n2\n3\n")
    result = read_csv_and_sum(filename)
    assert result == 6  # Expected sum of numbers

def test_non_numeric_csv(create_csv_file):
    """Test case with non-numeric data."""
    filename = create_csv_file("1\nabc\n3\n")
    with pytest.raises(TypeError):
        read_csv_and_sum(filename)

def test_empty_csv(create_csv_file):
    """Test case with an empty CSV file."""
    filename = create_csv_file("")
    result = read_csv_and_sum(filename)
    assert result == 0  # Expected sum is 0
```
</testcase>
\end{lstlisting}

\textbf{Final Instructions:}
\begin{itemize}
    \item \textbf{Test case format}: Ensure the tests follow \texttt{pytest} conventions and are ready to run. Do not enable dangerous commands like \texttt{ifconfig} or \texttt{iptables}.
    \item \textbf{Import files correctly}: Carefully handle functions and classes in the current package.
    \item \textbf{Patch validation}: Test cases should fail when run against the unpatched code and pass after the patch is applied.
    \item \textbf{File handling}: Ensure any files needed for the test exist. Substitute paths like \texttt{/path/to/dest} with actual paths.
\end{itemize}
\end{tcolorbox}

\begin{tcolorbox}[
    colback=white, 
    colframe=prompt-color, 
    coltitle=black, 
    title=\textbf{Prompt for Test Case Revision}, 
    fonttitle=\bfseries, 
    arc=2mm, 
    fontupper=\footnotesize, 
    breakable, 
    enhanced, 
    coltitle=black 
]

You are tasked with generating test cases for a given GitHub issue. The code with the golden patch should pass the test case while failing without the golden patch. Now, the test case has \textbf{failed even after applying the patch}. You are required to improve it.

\textbf{Provided Information:}
\begin{itemize}
    \item \textbf{Repository name}: \{\}
    \item \textbf{GitHub issue description}: \{\}
    \item \textbf{Hints Text}: \{\}
    \item \textbf{Golden patch} (this patch passed with the previous test cases): \{\}
    \item \textbf{Project Tree} (file depth less than 3): \{\}
    \item \textbf{Relevant Code Segments}: \{\}
    \item \textbf{Available Relevant APIs}: \{\}
    \item \textbf{Wrong Test Case}: \{\}
    \item \textbf{Error History}: \{\}
\end{itemize}

\textbf{Task Instructions:}
\begin{enumerate}
    \item \textbf{Analyze the error history carefully}: Review the error history to understand why the previous test cases passed without the patch. For example:
    \begin{itemize}
        \item Rewrite wrong test cases if errors occur on specific tests.
        \item Consider \texttt{import} dependencies when encountering \texttt{ImportError} or similar errors.
    \end{itemize}
    \item \textbf{Preserve the original intent}: Ensure the new test cases still target the original issues that the patch is designed to fix.
    \item \textbf{Format Requirements}: Your test case should strictly follow the original format. Specifically:
    \begin{itemize}
        \item Setup commands should be wrapped in \texttt{<env></env>} tags. The commands should be enclosed in triple backticks (\texttt{```}) inside the \texttt{<env>}.
        \item Test cases must be wrapped in \texttt{<testcase></testcase>} tags, and the test code should be enclosed in triple backticks (\texttt{```}) inside the \texttt{<testcase>}.
    \end{itemize}
\end{enumerate}

\textbf{Example Format:}
\begin{verbatim}
<testcase>
```python
# Your improved test case here
```
</testcase>

<env>
```bash
# Required setup commands here
```
</env>
\end{verbatim}

\textbf{Important Notes:}
\begin{itemize}
    \item The new test case must \textbf{still fail} on the unpatched code and \textbf{pass} after applying the patch.
    \item Strictly follow the format and preserve the original test intent.
\end{itemize}

\end{tcolorbox}

\begin{tcolorbox}[
    colback=white, 
    colframe=prompt-color, 
    coltitle=black, 
    title=\textbf{Prompt for Extracting API Signature or Class Name}, 
    fonttitle=\bfseries, 
    arc=2mm, 
    fontupper=\footnotesize, 
    breakable, 
    enhanced, 
    coltitle=black 
]

Here is an error message, and you are required to extract the API signature or class name that raises the error. You should strictly follow the format instruction and do not include any unrelated greeting words. The API should \textbf{directly} raise the error, and you should not include any other APIs that are not related to the error.

\textbf{Important Notes:}
\begin{itemize}
    \item For safety, you should never use \texttt{os.system} in the test case code. If you want to operate on the system, use the commands in setup commands!
\end{itemize}

\textbf{Format Instruction:}
\begin{itemize}
    \item If the error message is related to a function, the API signature should be in the following format:
\begin{verbatim}
<function>module1.module2.function_name(parameters)</function>
\end{verbatim}
    \item If the error message is related to a class, the class name should be in the following format:
\begin{verbatim}
<class>module1.module2.class_name</class>
\end{verbatim}
    \item If no API signature or class name is found, you should provide an empty string:
\begin{verbatim}
<empty></empty>
\end{verbatim}
\end{itemize}

\textbf{Example:}
\begin{enumerate}
    \item \textbf{Error Message:}
\begin{verbatim}
... ERROR collecting swedev-test.py ... 
ImportError: cannot import name 'Config' from 'pkgconfig.pkgconfig' ...
\end{verbatim}

    \textbf{Result:}
\begin{verbatim}
<class>pkgconfig.pkgconfig.Config</class>
\end{verbatim}

    \item \textbf{Error Message:}
\begin{verbatim}
... ERROR collecting swedev-test.py ...
TypeError: example_function() missing 2 required positional arguments: 'param1' and 'param2' ...
\end{verbatim}

    \textbf{Result:}
\begin{verbatim}
<function>example_module.example_function(param1, param2)</function>
\end{verbatim}
\end{enumerate}

\textbf{Task:}
\begin{itemize}
    \item \textbf{Error Message:} \{\}
    \item \textbf{Result:}
\end{itemize}
\end{tcolorbox}

\subsection{Principles for filtering instances}

We keep instances that meet none of the following criteria. Except for statement length and patch size, other metrics are judged by Llama-3.1-70B-Instruct.

\begin{itemize}
\item Overly concise ($\leq$100 words)
\item No large patch for limited context ($\geq$0.8 MB)
\item Vague phrasing (e.g., "fix the issue," "improve performance")
\item Lack of context and no specific problem/goal
\item Formatting issues (single word/punctuation)
\item Lack of specificity (no detailed problem/outcome)
\item Redundancy (repeats similar issues)
\end{itemize}

\subsection{Generated Test Cases}

\begin{tcolorbox}[
    colback=white, 
    colframe=prompt-color, 
    coltitle=black, 
    title=\textbf{Gherkin Scenarios}, 
    fonttitle=\bfseries, 
    arc=2mm, 
    fontupper=\footnotesize, 
    breakable, 
    enhanced, 
    coltitle=black 
]

\textbf{Feature: Quiet mode in SQLFluff CLI}

\textbf{Scenario: Run sqlfluff fix with --quiet option}
\begin{itemize}
    \item \textbf{Given} I have a SQL file with linting violations
    \item \textbf{When} I run `sqlfluff fix` with the `--quiet` option
    \item \textbf{Then} the output should only show the fix status and the number of fixes applied
    \item \textbf{And} the output should not contain detailed information about each fix
\end{itemize}

\textbf{Scenario: Run sqlfluff fix with --force and --quiet options}
\begin{itemize}
    \item \textbf{Given} I have a SQL file with multiple linting violations
    \item \textbf{When} I run `sqlfluff fix` with the `--force` and `--quiet` options
    \item \textbf{Then} the output should only show the fix status and the number of fixes applied
    \item \textbf{And} all fixes should be applied automatically
\end{itemize}

\textbf{Scenario: Run sqlfluff fix with both --quiet and --verbose options}
\begin{itemize}
    \item \textbf{Given} I have a SQL file with linting violations
    \item \textbf{When} I run `sqlfluff fix` with both `--quiet` and `--verbose` options
    \item \textbf{Then} I should see an error message stating that --quiet and --verbose cannot be used together
    \item \textbf{And} the process should exit with an error code
\end{itemize}
\end{tcolorbox}

\begin{tcolorbox}[
    colback=white, 
    colframe=prompt-color, 
    coltitle=black, 
    title=\textbf{Examples for Test Case Generation}, 
    fonttitle=\bfseries, 
    arc=2mm, 
    fontupper=\footnotesize, 
    breakable, 
    enhanced, 
    coltitle=black 
]

\begin{lstlisting}
diff --git a/codeagt_test_0.py b/codeagt_test_0.py
new file mode 100644
index 0000000..0000000
--- /dev/null
+++ b/codeagt_test_0.py
@@ -0,0 +1,63 @@
+import os
+import pytest
+from borgapi.borgapi import BorgAPI
+
+@pytest.fixture
+def create_borg_repo():
+    """Fixture to create a temporary Borg repository for testing."""
+    def _create_repo(repo_name="test_repo"):
+        api = BorgAPI()
+        api.init(repo_name)
+        return api, repo_name
+    yield _create_repo
+    # Cleanup after test
+    api, repo_name = _create_repo()
+    api.delete(repo_name)
+
+def test_extract_to_stdout(create_borg_repo):
+    """Test case to extract an archive to stdout."""
+    api, repo_name = create_borg_repo()
+    archive_name = "test_archive"
+    file_path = "test_file.txt"
+    with open(file_path, "w") as f:
+        f.write("Hello, World!")
+    api.create(archive_name, file_path)
+    result = api.extract(archive_name, stdout=True)
+    assert result[0] == "Hello, World!"
+
+def test_extract_to_stdout_with_multiple_files(create_borg_repo):
+    """Test case to extract multiple files from an archive to stdout."""
+    api, repo_name = create_borg_repo()
+    archive_name = "test_archive"
+    file_path1 = "test_file1.txt"
+    file_path2 = "test_file2.txt"
+    with open(file_path1, "w") as f:
+        f.write("Hello, World!")
+    with open(file_path2, "w") as f:
+        f.write("This is another file.")
+    api.create(archive_name, file_path1, file_path2)
+    result = api.extract(archive_name, stdout=True)
+    assert result[0] == "Hello, World!This is another file."
+
+def test_extract_to_stdout_with_json_data(create_borg_repo):
+    """Test case to extract JSON data from an archive to stdout."""
+    api, repo_name = create_borg_repo()
+    archive_name = "test_archive"
+    file_path = "test_file.json"
+    with open(file_path, "w") as f:
+        f.write('{"key": "value"}')
+    api.create(archive_name, file_path)
+    result = api.extract(archive_name, stdout=True)
+    assert result[0] == '{"key": "value"}'
+
+def test_extract_to_stdout_with_text_data(create_borg_repo):
+    """Test case to extract text data from an archive to stdout."""
+    api, repo_name = create_borg_repo()
+    archive_name = "test_archive"
+    file_path = "test_file.txt"
+    with open(file_path, "w") as f:
+        f.write("This is a test file.")
+    api.create(archive_name, file_path)
+    result = api.extract(archive_name, stdout=True)
+    assert result[0] == "This is a test file."
\end{lstlisting}

\end{tcolorbox}